%% file: main.tex
\documentclass[letterpaper, 10 pt, conference]{ieeeconf}  % Comment this line out if you need a4paper

\IEEEoverridecommandlockouts                              % This command is only needed if
\usepackage{times} % assumes new font selection scheme installed

\RequirePackage[loading]{tracefnt}

\PassOptionsToPackage{dvipsnames}{xcolor}   % added by yw for better colors
\usepackage{url}
\usepackage{amsmath}
\usepackage{algorithm}
\usepackage{algorithmic}
\usepackage[colorinlistoftodos,prependcaption,textsize=small]{todonotes} % added by yw
\usepackage{array,booktabs,threeparttable}
\usepackage{subcaption}
\usepackage{bbold}
\usepackage{adjustbox}
\usepackage{tikz}
\usepackage{pgfplots}
\usepackage{mathtools}
\DeclarePairedDelimiter{\ceil}{\lceil}{\rceil}
\pgfplotsset{compat=newest}
\usepackage{soul}

\makeatletter
\let\NAT@parse\undefined
\makeatother
\usepackage{hyperref}
\hypersetup{ % added by yw
  colorlinks,
  citecolor=PineGreen,
  linkcolor=red,
  urlcolor=blue
}
\usepackage[capitalise,noabbrev]{cleveref}  % added by yw

\graphicspath{{figures/}}

\newcommand{\argmax}[1]{\underset{#1}{\operatorname{arg}\,\operatorname{max}}\;}
\newcommand{\subsec}{\quad}   % bold-face topic divider

\title{\LARGE \bf
RoarNet: A Robust 3D Object Detection based on \\
RegiOn Approximation Refinement
}

\author{Kiwoo Shin$^{*}$$^\dagger$, Youngwook Paul Kwon$^{*}$$^\ddagger$ and Masayoshi Tomizuka$^\dagger$% <-this % stops a space
\thanks{The authors are with the department of Mechanical Engineering, University of California, Berkeley, CA 94720, US. {\tt\footnotesize \{kiwoo.shin, young, tomizuka\}@berkeley.edu}}% <-this % stops a space
\thanks{$^{*}$The authors contributed equally.}%
\thanks{$^\dagger$Mechanical Systems Control Lab, University of California, Berkeley, CA, USA.}%
\thanks{$^\ddagger$Phantom AI Inc., CA, USA.}%
}

\begin{document}
\maketitle
\thispagestyle{empty}
\pagestyle{empty}
%%%%%%%%%%%%%%%%%%%%%%%%%%%%%%%%%%%%%%%%%%%%%%%%%%%%%%%%%%%%%%%%%%%%%%%%%%%%%%%%
% \setcounter{footnote}{2}  % 1, 2 used for affiliation

\begin{abstract}
We present RoarNet, a new approach for 3D object detection from 2D image and 3D Lidar point clouds.
Based on two stage object detection framework~(\cite{girshick_fast_2015, ren_faster_2015}) with PointNet~\cite{qi_pointnet_2017} as our backbone network, we suggest several novel ideas to improve 3D object detection performance.

The first part of our method, RoarNet\_2D, estimates the 3D poses of objects from a monocular image, which approximates where to examine further, and derives multiple candidates that are geometrically feasible. This step significantly narrows down feasible 3D regions, which otherwise requires demanding processing of 3D point clouds in a huge search space.

Then the second part, RoarNet\_3D, takes the candidate regions and conducts in-depth inferences to conclude final poses in a recursive manner. Inspired by PointNet, RoarNet\_3D processes 3D point clouds directly without any loss of data, leading to precise detection.

We evaluate our method in KITTI, a 3D object detection benchmark. Our result shows that RoarNet has superior performance to state-of-the-art methods that are publicly available.
Remarkably, RoarNet also outperforms state-of-the-art methods even in settings where Lidar and camera are not time synchronized, which is practically important for actual driving environment.

RoarNet is implemented in Tensorflow~\cite{Abadi:2016:TSL:3026877.3026899} and publicly available with pretrained models.
\end{abstract}

\section{Introduction}

Recently, 3D object detection has become a crucial component in various fields such as mobile robots and autonomous vehicles. 3D object detection helps to understand the geometry of physical objects in 3D space that are important to predict future motion of objects. While there has been remarkable progress in the fields of image based 2D object detection and instance segmentation, 3D object detection is less explored in the literature. In this work, we study 3D object detection, which predicts 3D bounding boxes of objects from 2D image and 3D point clouds.

Most current 3D object detection systems transform 3D point clouds into 2D images by projecting point clouds onto ground plane~(Bird's Eye View) and/or depth map~(Perspective View). These systems apply convolutional neural networks on those transformed images to detect objects. Those approaches often rely on sensor-fusion methods to compensate the loss of data that occurs during projecting 3D point clouds onto lower dimensional 2D planes. However, these sensor-fusion based approaches require high quality synchronization between 2D camera sensor and 3D Lidar sensor, which itself is very challenging due to different sensor operating frequencies. When the synchronization condition breaks down, the performance of 3D object detection degrades significantly (\cref{sec:exp1}).
\begin{figure}
    \centering
    \begin{tabular}{c}
   \includegraphics[width=0.92\linewidth]{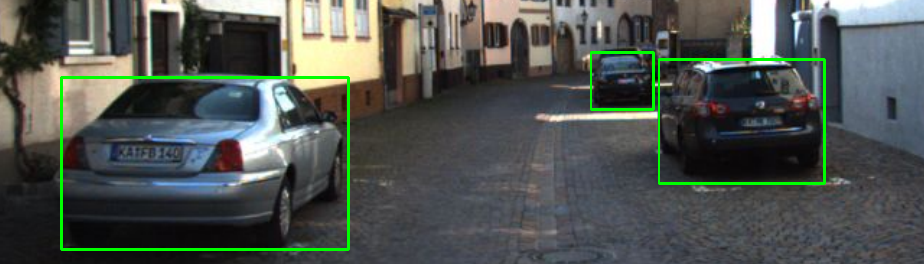}\\
   (a) geometric agreement search on 2D object detection\\
   \includegraphics[width=0.92\linewidth]{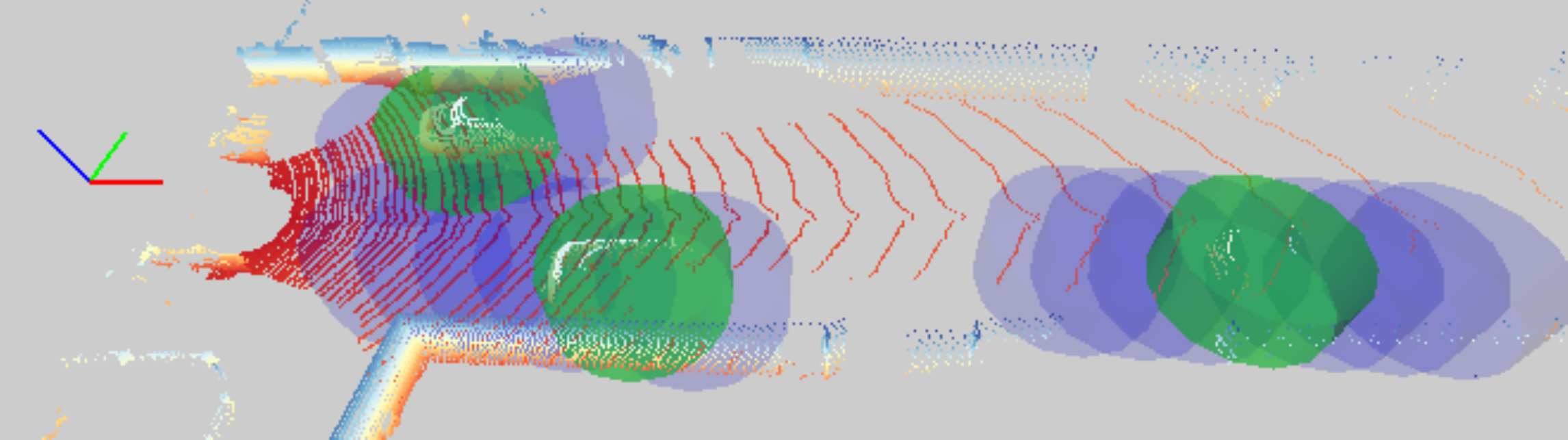}\\
   (b) 3D region proposals\\
   \includegraphics[width=0.92\linewidth]{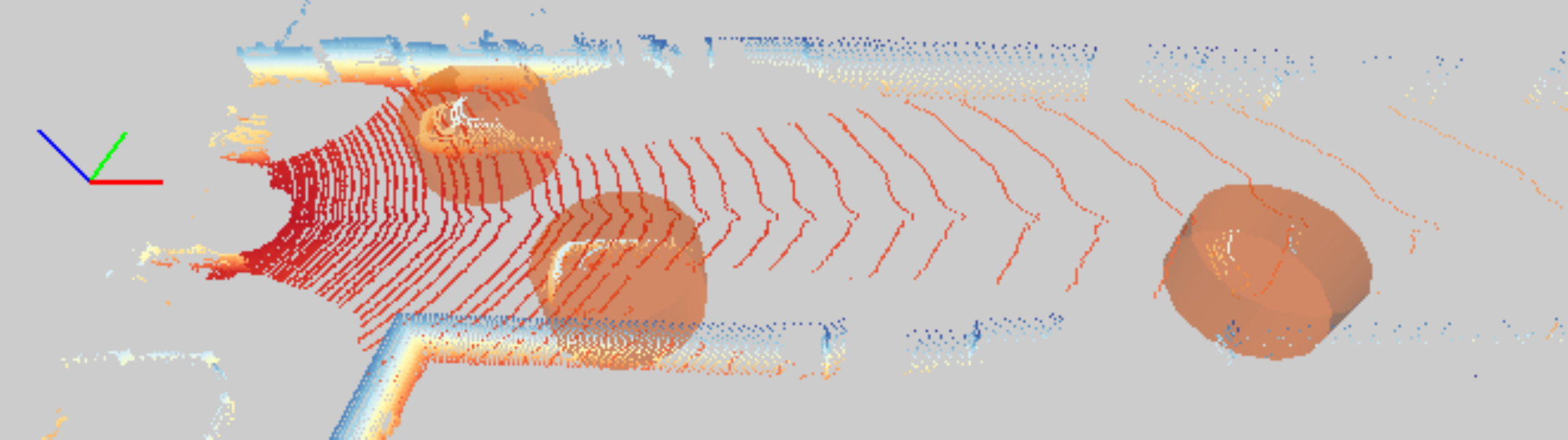}\\
   (c) 3D box regression\\
%     \includegraphics[width=0.43\linewidth]
%     {figures/Detect_step_IPM.pdf}&
%     \includegraphics[width=0.43\linewidth]{figures/Detect_step_Kazua_3D.pdf}\\
%     (b) 3D region proposals& (c) 3D box regression \\
    \includegraphics[width=0.92\linewidth]{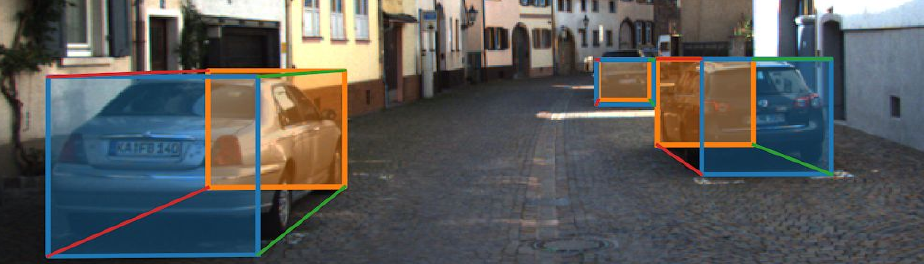}\\
   (d) resulting 3D bounding boxes
    \end{tabular}
    \caption{\textbf{Detection pipeline of RoarNet}. Our model (a) predicts region proposals in 3D space using geometric agreement search, (b) predicts objectness in each region proposal, (c) predicts 3D bounding boxes, (d) calculates IoU (Intersection over Union) between 2D detection and 3D detection.}
    \label{fig:3D_Detection_Step}
\end{figure}

Recently, \cite{qi_frustum_2018} predicts objects as 2D rectangular bounding boxes on the image plane and extend those boxes into 3D space along projection lines in the form of frustum. This makes it possible to filter the most of 3D point clouds out that are irrelevant to objects, and to process only those 3D point clouds that belong to objects directly without transforming the points to the 2D image plane. However, this approach is also sensitive to synchronization quality between sensors.

In this work, we propose a robust 3D detector, named \emph{RoarNet} (RegiOn Approximation Refinement Network), which helps to improve 3D object detection performance and reduce problems caused by sensor synchronization issue. RoarNet consists of two parts: RoarNet\_2D and RoarNet\_3D.

Inspired by geometric interpretation for monocular images in \cite{mousavian_3d_2017}, RoarNet\_2D estimates the 3D poses of objects from a monocular image and derives multiple candidate locations that are geometrically feasible, where the candidates are the input for RoarNet\_3D. This scheme significantly narrows down feasible 3D regions, which otherwise requires demanding processing of 3D point clouds in a huge search space (\cref{sec:roar_2d}).

% In this work, we suggest a novel approach for predicting region proposals, which helps to improve 3D object detection performance and reduce problem caused by sensor synchronization issue. Inspired by methods developed in the field of monocular pose estimation~\cite{chabot_deep_2017,mousavian_3d_2017,andriluka_monocular_2010,mehta2017,liang_deep_2018}, we formulate a new mathematical approach, named \emph{geometric agreement search} for predicting region proposals in 3D space. We describe this 2D to 3D region proposing step in a more detail in \cref{sec:roar_2d}.

Obtaining 3D region proposals predicted from 2D image,
RoarNet\_3D,
% we also suggest
a two-stage 3D object detector, gradually refines a search space making its training process efficient. The architecture of our model is analogous to standard two stage object detectors for 2D image such as Fast-RCNN and Faster-RCNN~\cite{girshick_fast_2015, ren_faster_2015}, and we adopt several modifications in order to make training of each stage easier (\cref{sec:roar_3d}).

The key difference compared to \cite{qi_frustum_2018} is that our model does not filter out point clouds by using 2D bounding box. Instead, our model takes the whole point clouds that are located inside region proposals which have the shape of standing cylinders. This leads to our model being more robust to sensor synchronization than state-of-the-art methods. We compare our method to other state-of-the-art 3D detection models in both synchronized and asynchronized conditions in \cref{sec:exp1}.

The detection pipeline of our model consists of three components as in \cref{fig:3D_Detection_Step}: (a) From a 2D image, our model predicts region proposals in 3D space. There can be multiple region proposals for a single detected object. (b) Using 3D point clouds sampled from the region proposals, we predict objectness in order to remove region proposals without foreground objects. At this step, we also predict the location of an object relative to given region proposals. We recursively use the relative location prediction as the center of region proposals for the next detection step. (c) Finally, our model predicts all coordinates for 3D bounding box regression including location, rotation and size. Practically, we repeat this step twice for better performance.
(d) To evaluate confidence of each detection, we calculate IoU (Intersection over Union) between the 2D detection and the 3D detection projected onto 2D image. The higher the correspondence between 2D detection and 3D detection is, the higher the confidence of detection is.

We evaluate our model on the 3D object detection task, provided by the KITTI benchmark. Our experiments show that RoarNet outperforms the state-of-the-art 3D object detection methods that are publicly available. We also evaluate our model in settings where camera and the Lidar are not time synchronized and the result shows that our model consistently performs better in these challenging settings.

All codes are implemented in Tensorflow and Cython and publicly available with several pretrained models. Additional materials are also available in \url{https://sites.google.com/berkeley.edu/roarnet}.

\section{Related Work}
\textbf{Monocular pose estimation\subsec} Due to the projection characteristics of camera sensors, monocular 3D pose estimation is very challenging. To overcome such difficulty, previous works often rely on domain knowledge or external data/information.
For example, human pose estimation applications were approached using a tracker (\cite{andriluka_monocular_2010}), through transferred learning from 2D and 3D datasets (\cite{mehta2017}) combined with the known skeleton topology of a human body.
In autonomous driving applications, \cite{chabot_deep_2017} trains a network to predict 36 control points per each vehicle that conveys 3D shape information. However, this method requires additionally annotating the auxiliary control point, which are very expensive to obtain. \cite{mousavian_3d_2017} proposes a novel method to predict physical dimensions (i.e, height, width, length in meters) and an orientation of vehicle without any additional data. Then, it can predict the location of object (i.e., $X, Y, Z$ in the world coordinate) by solving an over-constrained system of linear equations system. Since we find this method useful, we explore the method in more detail in \cref{sec:roar_2d} where we modify the method to be more computationally efficient and use it as our first building block for predicting region proposals in 3D space from a 2D image.

\textbf{3D point clouds processing\subsec} Since autonomous driving applications require very high level of accuracy in 3D pose estimation that monocular algorithms cannot provide, many algorithms using Lidar sensors are proposed. There are three popular representations to handle unstructured point clouds: (1) The first representation is using a 3D voxel grid \cite{zeng_wang_voting_2015,li_vehicle_2016-1,engelcke_vote3deep_2017-1}. In autonomous driving applications, however, sparse points clouds generally make voxel representation computationally redundant. (2) The second is to project an point cloud onto one or more 2D planes \cite{luo_fast_2018,simon_complex-yolo_2018,yang_pixor_2018}. These representations are usually compact and efficient, and can be treated as images. However, information loss by projection is inevitable. (3) The third one is to use the point clouds directly without any structured form. PointNet \cite{qi_pointnet_2017, qi_pointnet++_2017} showed how to digest point clouds directly for object classification and segmentation, and Frustum PointNet (F-PointNet) \cite{qi_frustum_2018} selects only necessary 3D points utilizing 2D detection results (i.e., 3D points within a frustum region that a camera position and a 2D bounding box make), and conducts detection using a PointNet scheme.

Similar to F-PointNet, advanced algorithms use both images and point clouds in a sensor fusion manner to enhance performance \cite{ku_joint_2017,liang_deep_2018}. Among these, F-PointNet and Aggregate View Object Detection (AVOD) \cite{ku_joint_2017} show the state-of-the-art performance on the public KITTI dataset leader board. RoarNet outperforms these methods in the standard 3D object detection, and our analysis shows that RoarNet shows better robustness in an even more general setting.

\begin{figure*}[t]
  \centering
  \includegraphics[width=0.97\linewidth]{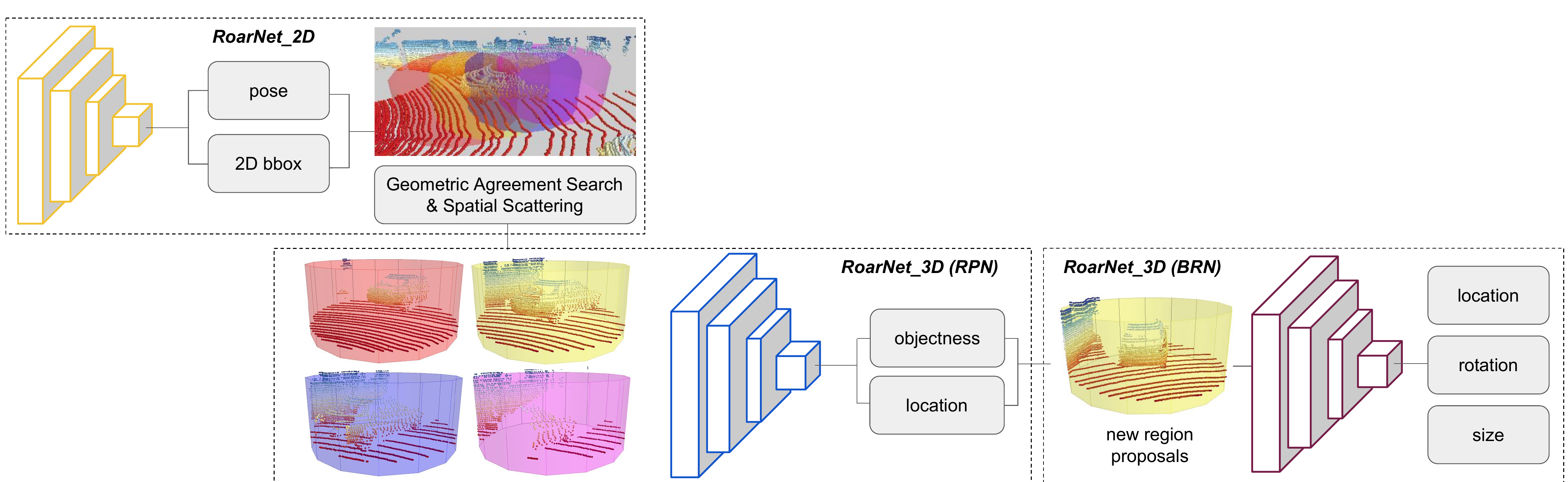}
  \caption{Architecture of RoarNet}
  \label{fig:3D_Detection_Network}
\end{figure*}

\section{Designing a RoarNet Detector}
The main idea behind RoarNet is to construct sequential networks that gradually refines a search space at each step in order to assign each network a simple task, and thus leads to efficient training and prediction.

% RoarNet is structured to utilize advantage of both 2D image and 3D point clouds, those contain supplementary information for 3D object detection. On the one hand, 2D image with high resolution provides reliable objectness information. On the other hand, point clouds provides exact spatial information of the objects. We design our model based on these characteristics.

\cref{fig:3D_Detection_Network} shows the architecture of RoarNet. The model first predicts the 2D bounding boxes and a 3D poses of objects from a 2D image. For each 2D object detection, geometric agreement search is applied to predict the location of object in 3D space.
% The inverse projection method is essentially based on a monocular pose estimation algorithm and we discuss a mathematical formulation behind the inverse projection method in a more detail in Sec. \cref{roar_2d}.
Centered on each location prediction, we set region proposal which has a shape of standing cylinder. Taking the prediction error in bounding box and pose into account, there can be multiple region proposals for a single object.

% our model sets sampling regions\footnote{In this work, we define sampling region as cylinder shaped manifold, since this is compatible with two key concepts discussed in section \cref{kazua_3d}}(figure AAA) which are centered on each approximated location from previous step.
Each region proposal is responsible for detecting a single object. Taking the point clouds sampled from each region proposal as input, our model predicts the location of an object relative to the center of region proposal, which recursively serves for setting new region proposals for the next step. Our model also predicts objectness score which reflects the probability of an object being inside the region proposal. Only those proposals with high objectness scores are considered at the next step.

At a final step, the model sets new region proposals at previously predicted locations.
% Assuming that prediction becomes more accurate at each step, the size of new region proposals is smaller than the previous step.
Our model predicts all coordinates required for 3D bounding box regression including location, rotation, and size of the objects.
% , but objectness score. This means that objectness score is determined only once from the previous step.
For practical reason, we observe that repeating this step more than once gives better detection performance.
% \begin{figure}[h]
%     \centering
%     \begin{tabular}{cc}
%     \includegraphics[width=0.4\linewidth]{figures/center_angle_before.pdf}&
%     \includegraphics[width=0.4\linewidth]{figures/center_angle_after.pdf}\\
%     (a) prediction at center angle&(b) actual prediction
%     \end{tabular}
%     \caption{Explain how center angle is applied: Figures and explanation must be modified soon.}
%     \label{fig:Center_Angle}
% \end{figure}

% To calculate detection score for each predicted 3D bounding box, we multiply three terms: the 2D object detection score, the probability of objectness and the intersection over union (IoU) between the bounding boxes at 2D image plane, one directly from 2D object detector and the other one which is projected from 3D bounding box into 2D image plane.
% \begin{equation}
% \text{score(3D)} = \text{score(2D)} * \text{Pr(Object)} * \text{IoU}^{\text{2D}}_{\text{3D}}
% \end{equation}

In \cref{sec:roar_2d}, we explain RoarNet\_2D that bridges image based 2D object detection to point clouds based 3D object detection. In \cref{sec:roar_3d}, we describe RoarNet\_3D, which predicts 3D bounding box using point clouds.

\subsection{RoarNet\_2D} \label{sec:roar_2d}
\textbf{Geometric agreement search\subsec}
For our initial seeds of 3D region proposals, we utilize a method suggested by \cite{mousavian_3d_2017} for monocular pose estimation, which we call \emph{geometric agreement search}:
% Although not much work has been done in monocular pose estimation, one of the recent success was proposed by \cite{mousavian_3d_2017}.
Given that the 3D pose of an object can be represented by seven degrees of freedom (localization in the camera coordinate $X, Y, Z$, physical dimensions of width, height and length $W, H, L$, and heading angle $\Theta$), a 2D bounding box window and the projection of its 3D pose (i.e., 3D box formed by $X, Y, Z, W, H, L, \Theta$ and camera projection matrix $P$) should agree. \cite{mousavian_3d_2017} showed that (1) a network can regress $\{W, H, L, \Theta\}$ per object, (2) there are only finite number of possible combinatorial configurations that a 3D box can locate to tightly fit a given 2D box, and (3) at each configuration, translation $X, Y, Z$ can be solved from known (regressed) $W, H, L, \Theta$ using an over-constrained system of linear equations. Then, the best configuration that minimizes projection error is selected.

More formally, for an object, let $b_{2D}$ be its 2D bounding box (from a 2D detector). At each configuration $c$, one can calculate a 3D bounding box candidate $b_{3D}^{c}$ as
\begin{align}
b_{3D}^{c} = B(W,H,L,\Theta; c, b_{2D}) \label{eq:lin}
\end{align}
where $B$ is the over-constrained linear equation system aforementioned.
The best configuration $c^{*}$ can be obtained by checking the agreement between $b_{2D}$ and the projection of 3D box $b_{3D}^{c}$.
\begin{align}
b_{PROJ}^{c} &= T(b_{3D}^{c}; P) \\
c^{*} &= \argmax{c\in C} \text{IoU}(b_{2D}, b_{PROJ}^{c})
\end{align}
where $T$ is projective transformation onto the image coordinate, $\text{IoU}$ is a widely-used intersection-over-union measure, and $C$ is the finate configuration set.\footnote{We refer \cite{mousavian_3d_2017} for the details about the configuration set $C$, and the over-constrained system of linear equations $B$.}

One drawback of \cite{mousavian_3d_2017} is that the $\{W,H,L,\Theta\}$ inference and inverse projection process should be done after running a separate 2D object detection and should be conducted for each detected vehicle. In other words, when an image includes $k$ objects, there should $k$-time computation of the network.

\begin{figure}
	\centering
	\begin{subfigure}{0.47\linewidth}
		\centering
		\includegraphics[height=4.2cm]{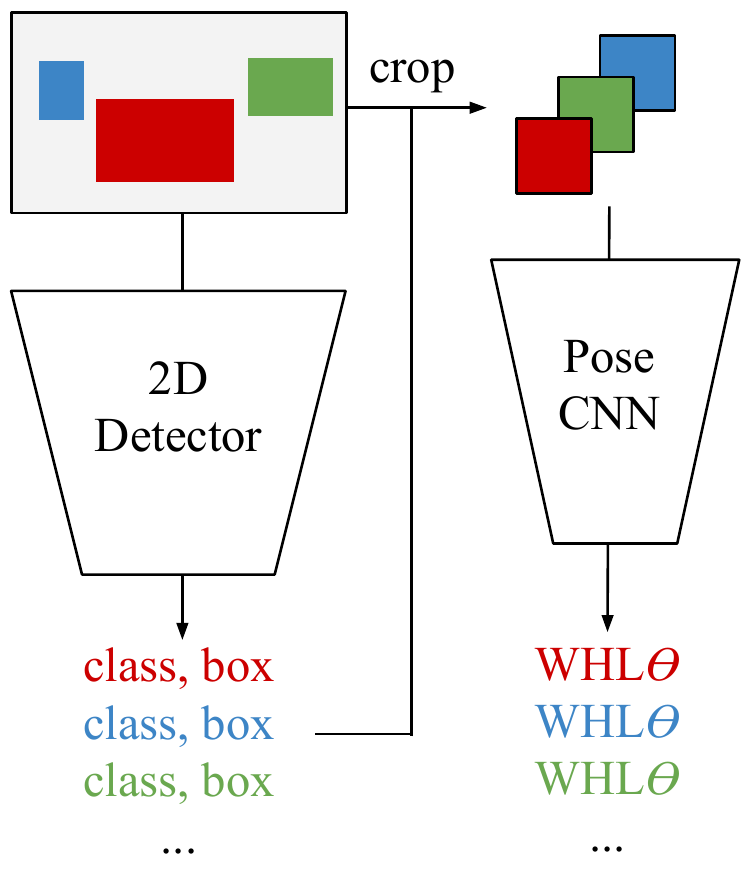}
		\caption{Architecture in \cite{mousavian_3d_2017}}\label{fig:mousavian}
	\end{subfigure}
    \hfill
    \begin{subfigure}{0.48\linewidth}
		\centering
        \includegraphics[height=4.2cm]{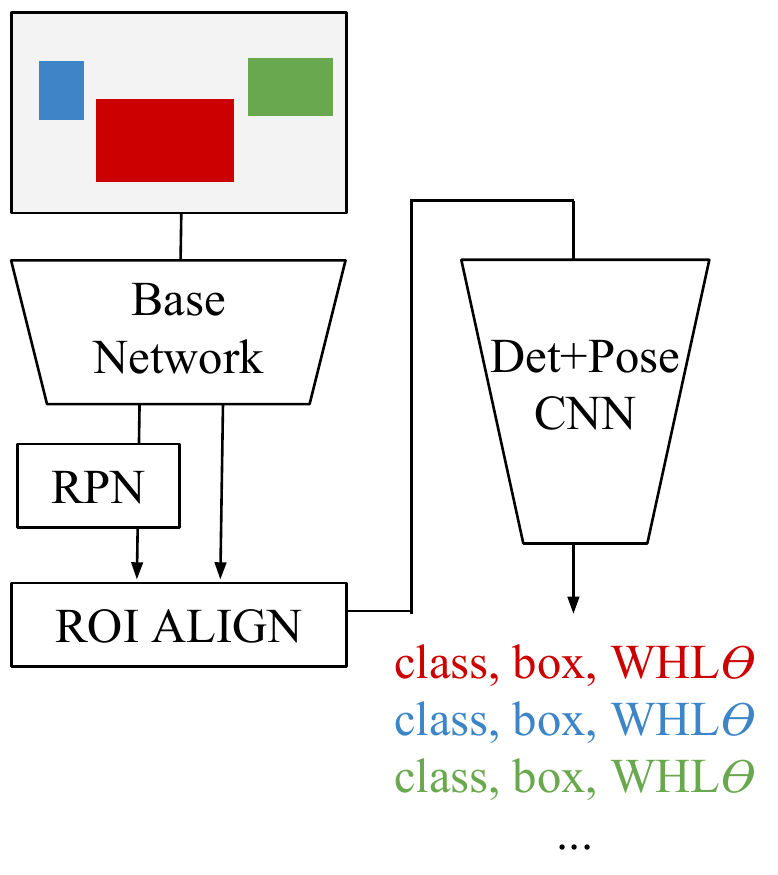}
%         \raisebox{.05\textwidth}{
% 		\includegraphics[width=\linewidth]{roar_2d_b.pdf}}
		\caption{RoarNet\_2D architecture}\label{fig:gas}
	\end{subfigure}
\caption{Architecture of RoarNet\_2D \label{fig:roar2d}}
\end{figure}

Aiming better computation efficiency, we build an unified network that combines the 2D object detection and $\{W,H,L,\Theta\}$ inference as illustrated in \cref{fig:gas}. In other words, the 2D bounding boxes and $\{W,H,L,\Theta\}$s of $k$ objects can be inferred with only one forward calculation of the unified network.

\textbf{Spatial scattering\subsec}
Note that the role of RoarNet\_2D, as a 3D region proposer, is to provide proposals of higher recall.
Since the monocular pose estimation suffers from limited accuracy, it is necessary to scatter our initial monocular pose estimation in order to increase the number of feasible pose candidates, and therefore, increase recall: For each object (i.e., its bounding box $b_{2D}$, regressed pose $XYZWHL\Theta$, and the best configuration $c^*$),
we first set a \emph{scattering range} by considering two extreme cases where the true physical size could actually be $1-s$ times smaller and $1+s$ times larger than the regressed size $WHL$ ($0<s<1$), which results in differently located 3D boxes by \cref{eq:lin}:
\begin{align*}
{b_{3D}^{c^*}}^{small} &= B((1-s)W,(1-s)H,(1-s)L,\Theta; c^{*}, b_{2D})\\
{b_{3D}^{c^*}}^{large} &= B((1+s)W,(1+s)H,(1+s)L,\Theta; c^{*}, b_{2D}).
\end{align*}
Recall that \cref{eq:lin} means the geometric constraint that the projection of the 3D box of an object should match with its 2D box, i.e., for the same 2D bounding box, smaller 3D boxes result in closer locations to the camera origin. Given these two extreme boxes, we divide the line of their two center points, $p1$ and $p2$, into an equal stride distance $m$. RoarNet\_2D detector finally provides $\ceil[\big]{\|p_1-p_2\|/m}$ 3D points per object for RoarNet\_3D to start.\footnote{$s=.5, m=1.6$ for experiments; $s=.2, m=1.25$ for \cref{fig:roar2d}.}

\begin{figure}
	\centering
	\begin{subfigure}[t]{\linewidth}
		\centering
		\includegraphics[width=\linewidth]{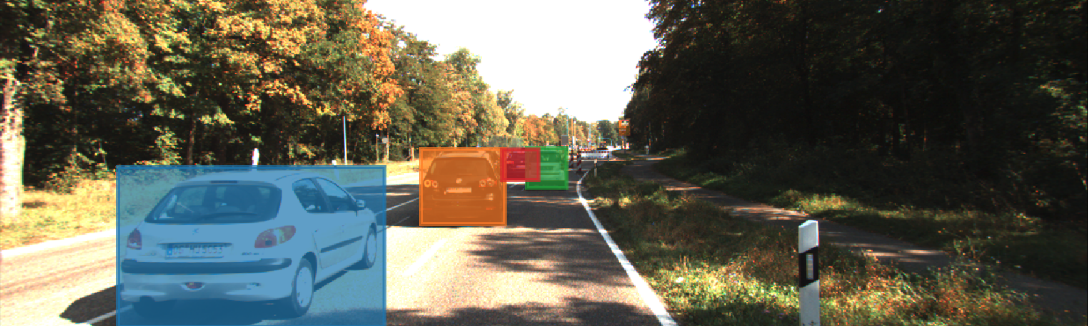}
		\caption{2D detection}\label{fig:inv_2d}
	\end{subfigure}
    \par\bigskip
    \begin{subfigure}[t]{\linewidth}
		\centering
		\input{roar2d.tex}
        \vspace{-0.5cm}
		\caption{Geometric agreement search and spatial scattering}
        \label{fig:inv_bv}
	\end{subfigure}
\caption{\textbf{RoarNet\_2D}. An unified architecture detects 2D bounding boxes and 3D poses illustrated as color-filled boxes in (a) and (b), respectively. For each object, two extreme cases are shown as non-filled boxes, and final equally-spaced candidate locations as colored dots in (b). All calculations are derived in 3D space despite bird's eye view (i.e., $XZ$ plane) visualization.\label{fig:inverse}}
\end{figure}

We visualize the process of RoarNet\_2D detector in \cref{fig:inverse}. RoarNet\_2D detector predicts 2D bounding boxes (\cref{fig:inv_2d}) as well as their physical sizes $WHL$ and heading angles $\Theta$, which lead to calculate their positions $XYZ$ (color-filled boxes in \cref{fig:inv_bv}). For each object, we consider two extreme deviations (non-filled boxes in \cref{fig:inv_bv}), and collect the uniform linear subdivision between the center points of the extreme poses (colored dots in \cref{fig:inv_bv}).

Note that the geometric agreement search and spatial scattering scheme significantly narrows down feasible 3D regions into a few linear regions, which otherwise requires a huge search space. Moreover, by virtue of geometric agreement constraints, our resulting proposals natively distribute (1) along the projection rays of the camera, and (2) in larger areas for more challenging further objects without bells and whistles.

\newlength\figureheight
\newlength\figurewidth

\subsection{RoarNet\_3D}\label{sec:roar_3d}
\begin{figure}[h]
    \centering
    \begin{tabular}{c}
    \includegraphics[width=0.84\linewidth]{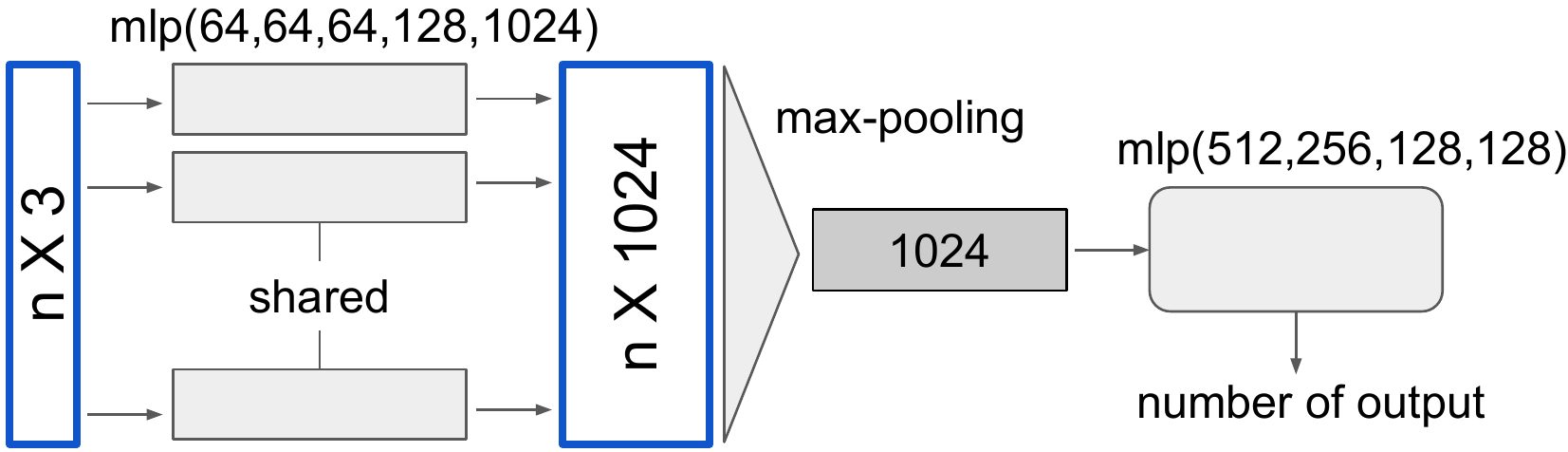}
    \end{tabular}
    \caption{\textbf{Our backbone network} is a simplified version of PointNet without T-Net in the original paper~\cite{qi_pointnet_2017}.}
    \label{fig:pointnet}
\end{figure}

\textbf{Network architecture\subsec}
The RoarNet\_3D is designed to predict a 3D bounding box that optimally fits for a given object by using point clouds. While building RoarNet\_3D as a two-stage object detector, the backbone network is inspired by the PointNet\cite{qi_pointnet_2017}, which uses max-pooling layers in the middle to get a global feature directly from unstructured point clouds. For more details, we refer readers to \cite{qi_pointnet_2017,qi_frustum_2018,qi_pointnet++_2017}. In this work, we use a simplified version of PointNet shown in \cref{fig:pointnet}.

RoarNet\_3D consists of two networks, called RPN (region proposal network) and BRN (box regression network), those have same structure except for the number of output as shown in \cref{fig:pointnet} and \cref{tab:output_kazua3d}.

\begin{table}[h]
\begin{center}
\begin{tabular}{@{}>{\centering}m{3.5cm}cc@{}}
\toprule
Number of Outputs & RPN & BRN\\
\midrule
location & 3 & 3\\
rotation  & 0 & 2*$N_{R}$\\
size & 0 & 4*$N_{C}$\\
objectness & 1 & 0\\
\bottomrule
\end{tabular}
\end{center}
\caption{Number of output at each network}
\label{tab:output_kazua3d}
\end{table}

The location is predicted by 3 coordinates $(t_{x}, t_{y}, t_{z})$ for (x, y, z) directions which is relative to center of region proposals. If a center of region proposal is offset from the origin by $(c_{x}, c_{y}, c_{z})$, then the location prediction corresponds to:
\begin{align}
\label{eqn:eqlabel}
\begin{split}
 x &= c_{x} + \text{2}* ( \sigma(t_{x}) \text{ - 0.5}) * m_{x} ,
\\
 y &= c_{y} + \text{2}* ( \sigma(t_{y}) \text{ - 0.5}) * m_{y} ,
\\
 z &= c_{z} + \text{2}* ( \sigma(t_{z}) \text{ - 0.5}) * m_{z}
\end{split}
\end{align}
We constrain the location prediction be bounded by $(m_{x}, m_{y}, m_{z})$ from center of region proposal.

The rotation angle is predicted by 2*$N_{R}$ coordinates ($t_{r\_\text{cls}(i)}, t_{r\_\text{reg}(i)})_{i=1}^{N_{R}}$ which is a hybrid formulation of $<$cls+reg$>$ structure. We equally divide [0, pi) to $N_{R}$ bins.

The size is predicted by 4*$N_{C}$ coordinates, $(t_{size\_cls(i)}, t_{h(i)}, t_{w(i)}, t_{l(i)})_{i=1}^{N_{C}}$ which is also a hybrid formulation of $<$cls+reg$>$ structure. We use K-Means method to get $N_{C}$ clusters.

The objectness is predicted by the output $t_{o}$ which reflects the probability of object or not object for each region proposal. We use sigmoid function to bound its value in a range of [0.0, 1.0$)$.
\subsection{Training and prediction}
During training each network, we optimize the following multi-task loss for RPN and BRN:
\begin{align}
\label{eqn:loss}
\begin{split}
 L_{\text{RPN}} &= \lambda_{\text{obj}} * L_{\text{obj}} + \mathbb{1}^{\text{obj}} [L_{\text{loc}}],
\\
 L_{\text{BRN}} &= \mathbb{1}^{\text{3D IoU} < 0.8} \Bigl[ L_{\text{loc}} + L_{\text{rot-cls}} + \mathbb{1}^{\text{rot-cls}}[L_{\text{rot-reg}}]
 \\
 & \qquad \qquad \qquad \quad+ L_{\text{size-cls}} + \mathbb{1}^{\text{size-cls}}[L_{\text{size-reg}}] \Bigr]
\end{split}
\end{align}

$L_{\text{loc}}, L_{\text{rot-reg}}$, and $L_{\text{size-reg}}$ are regression loss for location, rotation and size, which are represented as huber loss. $L_{\text{obj}}, L_{\text{rot-cls}}$, and $L_{\text{size-cls}}$ are classification loss for objectness, rotation and size, which are represented as cross-entropy loss.  $\mathbb{1}^{\text{obj}}$ denotes if objectness is true for a given region proposal. $\mathbb{1}^{\text{3D IoU}<0.8}$ is used for improving prediction performance for more general cases.

We down-sample point clouds with resolution of 0.1m for each axis. At each region proposal, we randomly sample 256 point clouds for training and 512 point clouds for prediction.

We train each network with batch of 512 for 500k iterations. Learning rate is 5e-3 for initial 100k and 5e-4 for rest of steps. It takes about two days for training each network with Titan X (not pascal).

Non-maximal suppression (NMS) is used to reduce redundant prediction at testing. We apply NMS on bird's eye view boxes with threshold of 0.05 to remove overlapping objects.

\section{Experiments}

\textbf{Dataset\subsec} We conduct our experiments in KITTI dataset, the 3D object detection benchmark. It provides synchronized 2D images and 3D LiDAR point clouds with annotations for car, pedestrian, and cyclist class. In this work, we focus on car class which has most training examples. The detection results are evaluated based on three difficulty levels: easy, moderate, and hard and we evaluate on moderate level, a standard metric for performance evaluation. 3D object detection performance is evaluated at 0.7 IoU threshold. Following \cite{qi_frustum_2018,ku_joint_2017,cai16mscnn}, we split training set into {\tt train} set of 3,717 frames and {\tt val} set of 3,769 frames such that frames in each split belong to different video clips.

\subsection{Comparison of the 3D object detection performance \label{sec:exp1}}

\begin{table}[t]
\begin{center}
\begin{tabular}{@{}>{}m{3cm}ccc@{}}
\toprule
Method & Easy & Moderate & Hard\\
\midrule
MV3D~\cite{cvpr17chen} 				& 71.09 		& 62.35 	 & 55.12\\
VoxelNet~\cite{zhou2017voxelnet}   			& 77.47 	   & 65.11 		& 57.73\\
UberATG-ContFuse~\cite{liang_deep_2018}  & 82.54 	   & 66.22 		& 64.04\\
F-PointNet (v2)~\cite{qi_frustum_2018}  & 81.20 		& 70.39 	 & 62.19\\
AVOD (FPN)~\cite{ku_joint_2017} 		& 81.94 	    & 71.88 	& \textbf{66.38}\\
% \hline
Ours 				 & \textbf{83.71} 		& \textbf{73.04} 	 & 59.16\\
% \hline
\bottomrule
\end{tabular}
\end{center}
\caption{3D object detection performance publicly available on the KITTI \textit{test} set, with 3D IoU threshold of 0.7}
\label{tab:comp_performance}
\end{table}

\begin{figure}
	\centering
    \setlength\figureheight{5.0cm}
    \setlength\figurewidth{\linewidth}
    \input{mytikz.tex}
\caption{A comparison of the 3D object detection performance in where Lidar and camera are not time synchronized. \label{fig:comparisons}}
\end{figure}
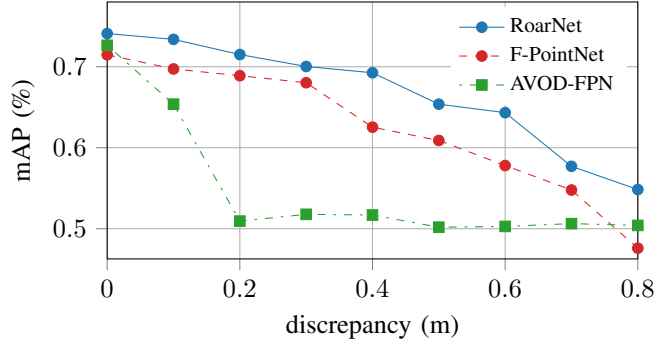

\textbf{Experiment settings\subsec} We evaluate our method in two settings. First, we evaluate our method in the original KITTI evaluation setting where the Lidar and the camera are well-synchronized each frame. This is a standard metric for ranking in KITTI benchmark leaderboard. Second, we evaluate our method in a more general case where the two sensors are not synchronized.
To simulate such case, we randomly translate the whole point clouds and re-generate ground truth labels according to the amount of translation of point clouds. This means that we regard the Lidar as the primary sensor. We constrain the translation of point clouds within 0.8m for x, y axis (i.e., parallel to the ground plane) and 0.2m for z axis (i.e., orthogonal to the ground plane).

\textbf{Experiment results\subsec} First, we evaluate RoarNet in a setting where the Lidar and the camera are synchronized, and compare it to publicly available 3D object detection methods on the KITTI benchmark. \cref{tab:comp_performance} shows that RoarNet shows state-of-the-art performance for 3D object detection in both easy and moderate level metric.

Second, we compare RoarNet to the two state-of-the-art methods, AVOD (FPN) and F-PointNet (v1) in a setting where sensors are \textit{not} synchronized. Those methods are selected since the AVOD (FPN) is the best among sensor-fusion based methods~\cite{ku_joint_2017} and the F-PointNet (v1) is the best among methods that directly process 3D point clouds~\cite{qi_frustum_2018, zhou2017voxelnet}.
\footnote{We train all methods for car class only. All methods are trained and evaluated in same train/val split.}

\cref{fig:comparisons} shows that RoarNet performs better than two state-of-the-art methods when two sensors are not synchronized. When sensors are synchronized, all three methods show the recall of 82.5\%. When two sensors are a-synchronized by 0.8m, the recall of our model degrades to 72.5\%, while the recall of F-PointNet degrades to 67.5\% and the recall of AVOD~(FPN) degrades to 65\%.

\subsection{Region proposals analysis}
In this section, we analyze the effect of spatial scattering parameter $s$ and objectness threshold in RoarNet\_3D (RPN) for refining a search space, as shown in \cref{fig:rpn_analysis}.

The smaller the value $s$, the higher confidence we have on monocular pose estimation. However, only 26.3\% of objects are captured in region proposals when we predict the location of object directly from monocular pose estimation ($s=0$). As we increase $s$, more objects are captured in region proposals, but number of region proposals are also linearly increased, which becomes the bottleneck of our detection pipeline. Aiming high recall, we use $s=0.5$ in our implementation.

The search space is further refined by RoarNet\_3D (RPN). In our implementation, we use objectness threshold of 0.25, that gives 83.2\% of recall with less than two region proposals per ground truth object.

\begin{figure}[h]
    \centering
    \setlength\figureheight{4cm}
    \input{rpn_dual_axes.tex}
    \setlength\figureheight{4cm}
    \input{rpn_3dtikz.tex}
    \caption{The effect of spatial scattering parameter $s$ and objectness threshold}
    \label{fig:rpn_analysis}
\end{figure}
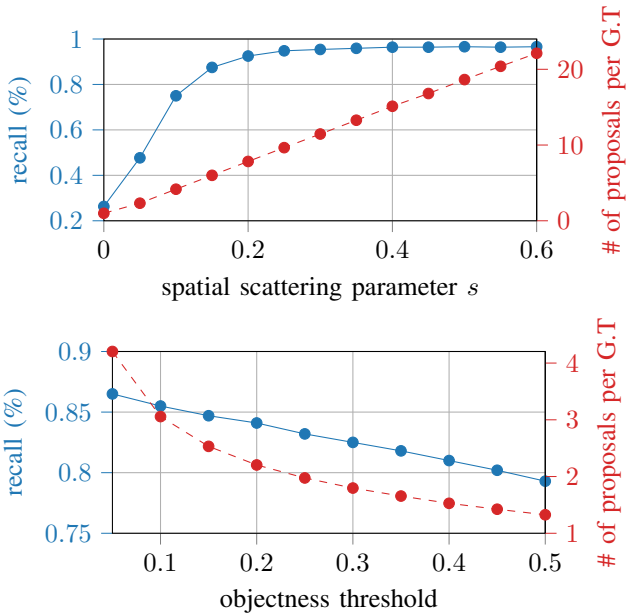
\begin{figure}[h]
    \centering
    \begin{tabular}{cc}
    \includegraphics[width=0.35\linewidth]{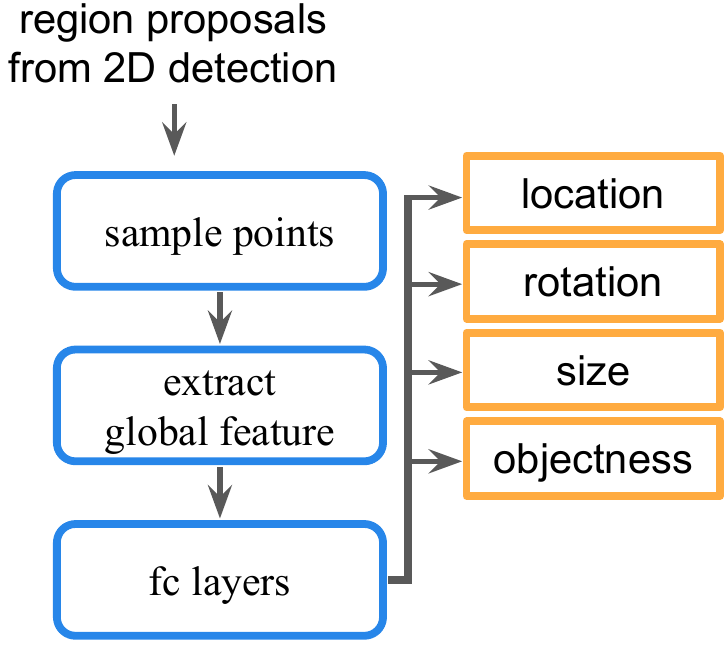}&
    \includegraphics[width=0.45\linewidth]{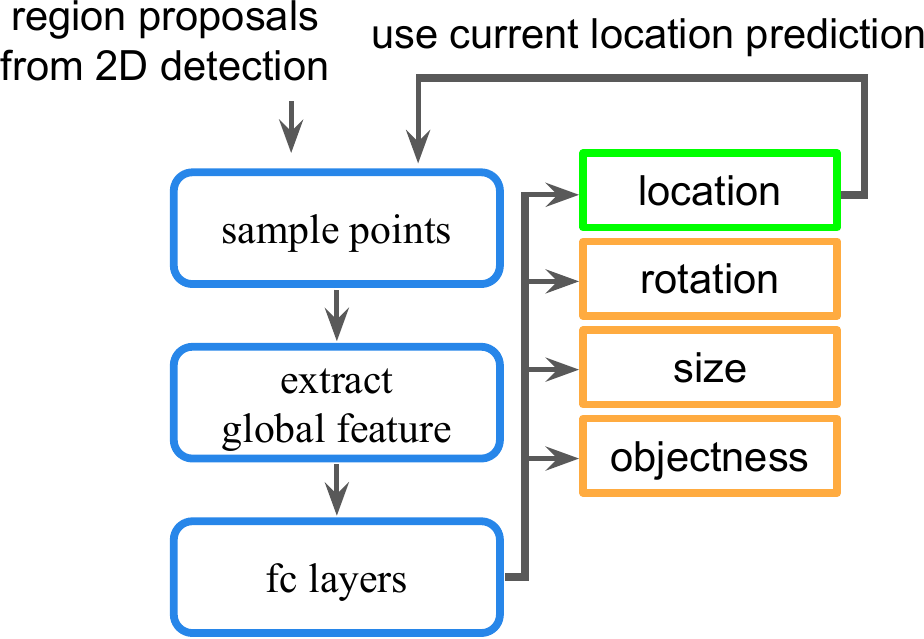}\\
    (a) single stage 3D detector&(b) run (a) twice\\
    \multicolumn{2}{c}{\includegraphics[width=0.80\linewidth]{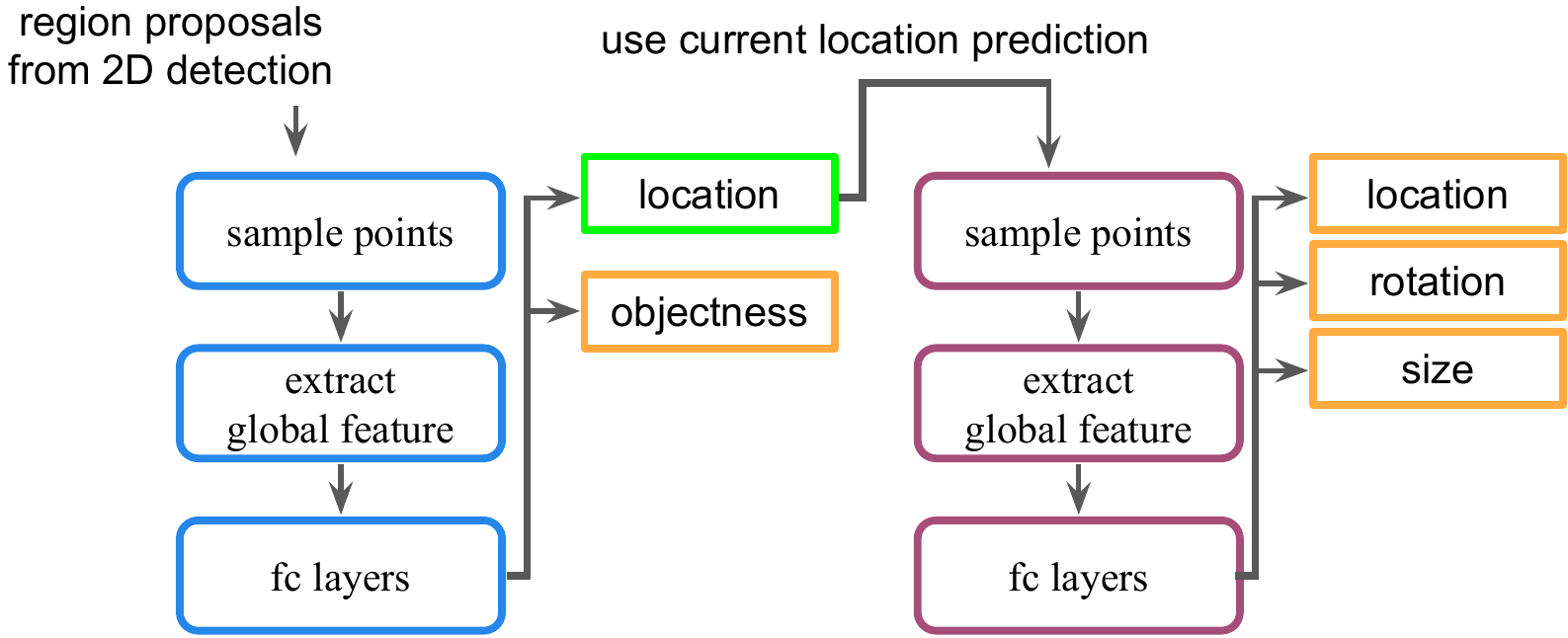}}\\
    \multicolumn{2}{c}{(c) our final model, RoarNet\_3D}\\
    \end{tabular}
    \caption{A detection pipeline of several network architectures}
    \label{fig:network_architecture}
\end{figure}

\subsection{Network design analysis}
In this section, we compare three network architectural designs shown in \cref{fig:network_architecture}.

\cref{fig:network_architecture}(a) represents a single stage 3D object detector, which predicts 3D bounding box along with objectness in a single step. This approach is inspired by YOLO detector~\cite{redmon2016you,redmon2017yolo9000}, which shows promising results in a 2D object detection. However, (a) shows the recall of 67.5\% and mAP of 54.3\%.

Without any further training step, we only modify the detection pipeline of (a) to use location predicted at current step as region proposals for the next step. This simple modification immediately improves the performance to 59.9\% with an increase of 5.6\% from (a).

This result inspires us to build our final model, RoarNet\_3D in \cref{fig:network_architecture}(c) that specializes each detection step to a specific task and remove redundant predictions. This modification leads significant performance improvement such that recall is 82.5\% and mAP is 74.02\%.

\section{Conclusion}

We have proposed RoarNet, a new approach for 3D object detection from an 2D image and 3D Lidar point clouds. RoarNet refines search space recursively at each step in order to make training and prediction efficient. We first estimate 3D poses from a monocular input image, and derives multiple geometrically feasible candidates nearby the initial estimates. We adopt a two-stage object detection framework to further refine search space effectively from 3D point clouds. Our model shows superior performance to state-of-the-art methods in KITTI, a 3D object detection benchmark. RoarNet outperforms even in the setting where Lidar and camera are not time synchronized, which is practically important results in order to extend current single frame based detection into video frame based detection in the future research.

\section*{ACKNOWLEDGMENT}
The work was in part supported by Berkeley Deep Drive. Kiwoo Shin is supported by Samsung Scholarship.

\newpage

\bibliographystyle{IEEEtran}
\bibliography{references,roar}

\end{document}

%% file: roar2d.tex
% This file was created by matplotlib2tikz v0.6.18.
\noindent\begin{tikzpicture}
\definecolor{color3}{rgb}{0.83921568627451,0.152941176470588,0.156862745098039}
\definecolor{color2}{rgb}{0.172549019607843,0.627450980392157,0.172549019607843}
\definecolor{color0}{rgb}{0.12156862745098,0.466666666666667,0.705882352941177}
\definecolor{color1}{rgb}{1,0.498039215686275,0.0549019607843137}

\begin{axis}[
axis equal image,
width=\textwidth+0.7cm,
tick align=outside,
tick pos=left,
x grid style={lightgray!92.02614379084967!black},
xlabel={Z},
xmajorgrids,
xmin=0, xmax=45,
y grid style={lightgray!92.02614379084967!black},
ymajorgrids,
ymin=-7, ymax=7,
font=\scriptsize,
xtick distance=5,
y dir=reverse
]
\path [draw=color0, fill=color0, opacity=0.5] (axis cs:5.7070871810893,-2.23209476169232)
--(axis cs:5.75450602807629,-3.77136454082476)
--(axis cs:9.1129128189107,-3.66790523830768)
--(axis cs:9.06549397192371,-2.12863545917524)
--cycle;

\path [draw=color0, line width=0.3mm, draw opacity=1.0, fill opacity=0] (axis cs:4.52435082473594,-1.78754156824973)
--(axis cs:4.56228590232554,-3.01895739155568)
--(axis cs:7.24901133499307,-2.93618994954201)
--(axis cs:7.21107625740347,-1.70477412623606)
--cycle;

\path [draw=color0, line width=0.3mm, draw opacity=1.0, fill opacity=0] (axis cs:7.07077036857353,-2.75941490749871)
--(axis cs:7.13004392730728,-4.68350213141425)
--(axis cs:11.3280524158503,-4.5541780032679)
--(axis cs:11.2687788571165,-2.63009077935235)
--cycle;

\path [draw=color1, fill=color1, opacity=0.5] (axis cs:12.8944094031476,-0.529807531675296)
--(axis cs:12.8289560380867,-2.19852436115902)
--(axis cs:17.2055905968524,-2.3701924683247)
--(axis cs:17.2710439619133,-0.701475638840982)
--cycle;

\path [draw=color1, line width=0.3mm, draw opacity=1.0, fill opacity=0] (axis cs:10.1527886843223,-0.430197105281032)
--(axis cs:10.1004259922736,-1.76517056886801)
--(axis cs:13.6017336392861,-1.90250505460056)
--(axis cs:13.6540963313348,-0.567531591013581)
--cycle;

\path [draw=color1, line width=0.3mm, draw opacity=1.0, fill opacity=0] (axis cs:15.8652079514443,-0.638563707272953)
--(axis cs:15.7833912451182,-2.72445974412761)
--(axis cs:21.2541844435753,-2.93904487808471)
--(axis cs:21.3360011499014,-0.853148841230061)
--cycle;

\path [draw=color2, fill=color2, opacity=0.5] (axis cs:23.9744150533986,1.40411186267243)
--(axis cs:23.9755856536634,-0.0658876712373659)
--(axis cs:26.2055849466014,-0.0641118626724309)
--(axis cs:26.2044143463366,1.40588767123737)
--cycle;

\path [draw=color2, line width=0.3mm, draw opacity=1.0, fill opacity=0] (axis cs:18.9852061455021,1.09511319581116)
--(axis cs:18.9861426257139,-0.0808864313166751)
--(axis cs:20.7701420600643,-0.0794657844647269)
--(axis cs:20.7692055798525,1.09653384266311)
--cycle;

\path [draw=color2, line width=0.3mm, draw opacity=1.0, fill opacity=0] (axis cs:29.6663975704104,1.74474952282143)
--(axis cs:29.6678608207413,-0.0927498945658115)
--(axis cs:32.4553599369138,-0.0905301338596426)
--(axis cs:32.4538966865829,1.7469692835276)
--cycle;

\path [draw=color3, fill=color3, opacity=0.5] (axis cs:31.4160792704677,0.171974175927813)
--(axis cs:31.5511799909196,-1.31188828243327)
--(axis cs:35.7239207295323,-0.931974175927813)
--(axis cs:35.5888200090804,0.551888282433271)
--cycle;

\path [draw=color3, line width=0.3mm, draw opacity=1.0, fill opacity=0] (axis cs:24.8724711739425,0.119900821493288)
--(axis cs:24.980551750304,-1.06718914519558)
--(axis cs:28.3187443411942,-0.763257859991213)
--(axis cs:28.2106637648327,0.423832106697654)
--cycle;

\path [draw=color3, line width=0.3mm, draw opacity=1.0, fill opacity=0] (axis cs:38.8647012914538,0.220964459774116)
--(axis cs:39.0335771920186,-1.63386361317724)
--(axis cs:44.2495031152845,-1.15897098004542)
--(axis cs:44.0806272147197,0.695857092905938)
--cycle;

\addplot [very thick, color1, opacity=0.5, forget plot]
table [row sep=\\]{%
9.1129128189107	-3.66790523830768 \\
9.06549397192371	-2.12863545917524 \\
};
\addplot [very thick, color1, opacity=0.5, forget plot]
table [row sep=\\]{%
7.24901133499307	-2.93618994954201 \\
7.21107625740347	-1.70477412623606 \\
};
\addplot [very thick, color1, opacity=0.5, forget plot]
table [row sep=\\]{%
11.3280524158503	-4.5541780032679 \\
11.2687788571165	-2.63009077935235 \\
};
\addplot [very thick, color1, opacity=0.5, forget plot]
table [row sep=\\]{%
17.2055905968524	-2.3701924683247 \\
17.2710439619133	-0.701475638840982 \\
};
\addplot [very thick, color1, opacity=0.5, forget plot]
table [row sep=\\]{%
13.6017336392861	-1.90250505460056 \\
13.6540963313348	-0.567531591013581 \\
};
\addplot [very thick, color1, opacity=0.5, forget plot]
table [row sep=\\]{%
21.2541844435753	-2.93904487808471 \\
21.3360011499014	-0.853148841230061 \\
};
\addplot [very thick, color1, opacity=0.5, forget plot]
table [row sep=\\]{%
26.2055849466014	-0.0641118626724309 \\
26.2044143463366	1.40588767123737 \\
};
\addplot [very thick, color1, opacity=0.5, forget plot]
table [row sep=\\]{%
20.7701420600643	-0.0794657844647269 \\
20.7692055798525	1.09653384266311 \\
};
\addplot [very thick, color1, opacity=0.5, forget plot]
table [row sep=\\]{%
32.4553599369138	-0.0905301338596426 \\
32.4538966865829	1.7469692835276 \\
};
\addplot [very thick, color1, opacity=0.5, forget plot]
table [row sep=\\]{%
35.7239207295323	-0.931974175927813 \\
35.5888200090804	0.551888282433271 \\
};
\addplot [very thick, color1, opacity=0.5, forget plot]
table [row sep=\\]{%
28.3187443411942	-0.763257859991213 \\
28.2106637648327	0.423832106697654 \\
};
\addplot [very thick, color1, opacity=0.5, forget plot]
table [row sep=\\]{%
44.2495031152845	-1.15897098004542 \\
44.0806272147197	0.695857092905938 \\
};
\addplot [semithick, color0, mark=*, mark size=2, mark options={solid}, only marks, forget plot]
table [row sep=\\]{%
5.8866810798645	-2.36186575889587 \\
7.54304623603821	-3.00933110713959 \\
9.19941139221191	-3.6567964553833 \\
};
\addplot [semithick, color1, mark=*, mark size=2, mark options={solid}, only marks, forget plot]
table [row sep=\\]{%
11.8772611618042	-1.1663510799408 \\
13.5478699207306	-1.32196438312531 \\
15.218478679657	-1.47757768630981 \\
16.8890874385834	-1.63319098949432 \\
18.5596961975098	-1.78880429267883 \\
};
\addplot [semithick, color2, mark=*, mark size=2, mark options={solid}, only marks, forget plot]
table [row sep=\\]{%
19.8776741027832	0.507823705673218 \\
21.4752747671945	0.5534359897886 \\
23.0728754316057	0.599048273903983 \\
24.670476096017	0.644660558019366 \\
26.2680767604283	0.690272842134748 \\
27.8656774248396	0.735885126250131 \\
29.4632780892508	0.781497410365513 \\
31.0608787536621	0.827109694480896 \\
};
\addplot [semithick, color3, mark=*, mark size=2, mark options={solid}, only marks, forget plot]
table [row sep=\\]{%
26.5956077575684	-0.321678519248962 \\
28.257996029324	-0.338047934903039 \\
29.9203843010796	-0.354417350557115 \\
31.5827725728353	-0.370786766211192 \\
33.2451608445909	-0.387156181865268 \\
34.9075491163466	-0.403525597519345 \\
36.5699373881022	-0.419895013173421 \\
38.2323256598579	-0.436264428827498 \\
39.8947139316135	-0.452633844481574 \\
41.5571022033691	-0.469003260135651 \\
};
\addplot [semithick, black, mark=triangle*, mark size=3, mark options={solid,rotate=90}, only marks, forget plot]
table [row sep=\\]{%
0	0 \\
};
\end{axis}

\end{tikzpicture}

%% file: mytikz.tex
% This file was created by matplotlib2tikz v0.6.18.
\begin{tikzpicture}

\definecolor{color2}{rgb}{0.172549019607843,0.627450980392157,0.172549019607843}
\definecolor{color1}{rgb}{0.83921568627451,0.152941176470588,0.156862745098039}
\definecolor{color0}{rgb}{0.12156862745098,0.466666666666667,0.705882352941177}

\begin{axis}[
height=\figureheight,
width=\figurewidth,
legend cell align={left},
legend entries={{RoarNet},{F-PointNet},{AVOD-FPN}},
legend style={draw=none,font=\footnotesize},
tick align=outside,
tick pos=left,
x grid style={lightgray!92.02614379084967!black},
xlabel={discrepancy (m)},
xmajorgrids,
xmin=0.0, xmax=0.8,
y grid style={lightgray!92.02614379084967!black},
ylabel={mAP (\%)},
ymajorgrids,
ymin=0.462742305, ymax=0.78
]
\addplot [color0, mark=*, mark size=2, mark options={solid}]
table [row sep=\\]{%
0	0.740967 \\
0.1	0.733839 \\
0.2	0.71508 \\
0.3	0.700341 \\
0.4	0.6926 \\
0.5	0.6538 \\
0.6	0.64332 \\
0.7	0.57706 \\
0.8	0.5484 \\
};
\addplot [color1, dashed, mark=*, mark size=2, mark options={solid,fill=none}]
table [row sep=\\]{%
0	0.714817 \\
0.1	0.697284 \\
0.2	0.68896347 \\
0.3	0.680298 \\
0.4	0.62539841 \\
0.5	0.609 \\
0.6	0.578 \\
0.7	0.5478 \\
0.8	0.4759911 \\
};
\addplot [color2, dash pattern=on 1pt off 3pt on 3pt off 3pt, mark=square*, mark size=2, mark options={solid}]
table [row sep=\\]{%
0	0.7265 \\
0.1	0.6538 \\
0.2	0.5094 \\
0.3	0.5178 \\
0.4	0.5169 \\
0.5	0.5019 \\
0.6	0.5029 \\
0.7	0.5063 \\
0.8	0.5041 \\
};
\end{axis}

\end{tikzpicture}

%% file: rpn_dual_axes.tex
% This file was created by matplotlib2tikz v0.6.18.
\begin{tikzpicture}

\definecolor{color2}{rgb}{0.172549019607843,0.627450980392157,0.172549019607843}
\definecolor{color1}{rgb}{0.83921568627451,0.152941176470588,0.156862745098039}
\definecolor{color0}{rgb}{0.12156862745098,0.466666666666667,0.705882352941177}

\begin{axis}[
% axis y line*=left,
height=\figureheight,
width=\linewidth-1.3cm,
% legend cell align={left},
% legend entries={{Recall},{F-PointNet},{AVOD-FPN}},
% legend style={draw=none,font=\footnotesize},
tick align=outside,
tick pos=left,
x grid style={lightgray!92.02614379084967!black},
xlabel={spatial scattering parameter $s$},
ylabel style=color0,
yticklabel style=color0,
ytick style=color0,
xmajorgrids,
xmin=0.00, xmax=0.6,
y grid style={lightgray!92.02614379084967!black},
ylabel={recall (\%)},
ymajorgrids,
ymin=0.20, ymax=1.0
]
\addplot [color0, mark=*, mark size=2, mark options={solid}]
table [row sep=\\]{%
0	0.263 \\
0.05 0.477 \\
0.1	0.750 \\
0.15 0.875 \\
0.2	0.925 \\
0.25 0.948 \\
0.3	0.954 \\
0.35 0.959 \\
0.4	0.964 \\
0.45 0.964 \\
0.5	0.966 \\
0.55 0.964 \\
0.6	0.966 \\
};
\end{axis}
\begin{axis}[
axis y line*=right,
% axis y line=none,
axis x line=none,
height=\figureheight,
width=\linewidth-1.3cm,
legend cell align={left},
legend style={draw=none,font=\footnotesize,at={(0.99,0.02)},anchor=south east},
tick align=outside,
% tick pos=left,
% x grid style={lightgray!92.02614379084967!black},
% xlabel={spatial scattering value $s$},
xmin=0.00, xmax=0.6,
ymin=0.0, ymax=24.0,  % <<- you need to adjust so that the # of tick grids should be the same to look good.
ylabel style=color1,
yticklabel style=color1,
ytick style=color1,
% y grid style={lightgray!92.02614379084967!black},
ylabel={\# of proposals per G.T},
% xmajorgrids,
% ymajorgrids
]
\addplot [color1, dashed, mark=*, mark size=2, mark options={solid,fill=none}]
table [row sep=\\]{%
0.0 0.988 \\
0.05  2.321\\
0.1	4.168 \\
0.15 6.000 \\
0.2	7.832 \\
0.25 9.654 \\
0.3	11.453 \\
0.35 13.276 \\
0.4	15.118 \\
0.45 16.803 \\
0.5	18.6399 \\
0.55 20.394 \\
0.6	22.117 \\
};
\end{axis}

\end{tikzpicture}

%% file: rpn_3dtikz.tex
% This file was created by matplotlib2tikz v0.6.18.
\begin{tikzpicture}

\definecolor{color2}{rgb}{0.172549019607843,0.627450980392157,0.172549019607843}
\definecolor{color1}{rgb}{0.83921568627451,0.152941176470588,0.156862745098039}
\definecolor{color0}{rgb}{0.12156862745098,0.466666666666667,0.705882352941177}

\begin{axis}[
% axis y line*=left,
height=\figureheight,
width=\linewidth-1.3cm,
% legend cell align={left},
% legend entries={{Recall},{F-PointNet},{AVOD-FPN}},
% legend style={draw=none,font=\footnotesize},
tick align=outside,
tick pos=left,
x grid style={lightgray!92.02614379084967!black},
xlabel={objectness threshold},
ylabel style=color0,
yticklabel style=color0,
ytick style=color0,
xmajorgrids,
xmin=0.05, xmax=0.5,
y grid style={lightgray!92.02614379084967!black},
ylabel={recall (\%)},
ymajorgrids,
ymin=0.75, ymax=0.9
]
\addplot [color0, mark=*, mark size=2, mark options={solid}]
table [row sep=\\]{%
0.05 0.865 \\
0.1	0.855 \\
0.15 0.847 \\
0.2	0.841 \\
0.25 0.832 \\
0.3	0.825 \\
0.35 0.818 \\
0.4	0.810 \\
0.45 0.802 \\
0.5	0.793 \\
};
\end{axis}
\begin{axis}[
axis y line*=right,
% axis y line=none,
axis x line=none,
height=\figureheight,
width=\linewidth-1.3cm,
legend cell align={left},
legend style={draw=none,font=\footnotesize,at={(0.99,0.02)},anchor=south east},
tick align=outside,
% tick pos=left,
% x grid style={lightgray!92.02614379084967!black},
% xlabel={spatial scattering value $s$},
xmin=0.05, xmax=0.5,
ymin=1.0, ymax=4.204,  % <<- you need to adjust so that the # of tick grids should be the same to look good.
ylabel style=color1,
yticklabel style=color1,
ytick style=color1,
% y grid style={lightgray!92.02614379084967!black},
ylabel={\# of proposals per G.T},
% xmajorgrids,
% ymajorgrids
]
\addplot [color1, dashed, mark=*, mark size=2, mark options={solid,fill=none}]
table [row sep=\\]{%
0.05  4.204\\
0.1	3.057 \\
0.15 2.532 \\
0.2	2.204 \\
0.25 1.975 \\
0.3	1.797 \\
0.35 1.656 \\
0.4	1.528 \\
0.45 1.424 \\
0.5	1.326 \\
};
\end{axis}

\end{tikzpicture}